\documentclass[letterpaper,twocolumn,10pt,anonymous]{article}
\usepackage{usenix-2020-09}

\usepackage{tikz}
\usepackage{amsmath}
\usepackage{enumitem}
\usepackage{filecontents}
\usepackage{xspace}
\usepackage{float}
\usepackage{booktabs}
\usepackage{graphicx}
\usepackage{subcaption}
\usepackage{amsmath}
\usepackage{balance}

\usepackage{amssymb}
\usepackage{makecell}
\usepackage{array}
\usepackage[ruled,linesnumbered,noend]{algorithm2e}
\usepackage{listings}
\usepackage{graphicx}
\usepackage{subcaption}
\usepackage{multirow}

\usepackage{changepage}
\usepackage[normalem]{ulem}

\usepackage{pifont}

\newcommand{\oursys}{\textsc{DynaTrain}\xspace}

\newcommand{\para}[1]{{\vspace{5pt} \bf \noindent #1 \hspace{5pt}}}

\newcommand{\squishlist}{
	\begin{list}{$\bullet$}
		{ \setlength{\itemsep}{0pt}      \setlength{\parsep}{3pt}
			\setlength{\topsep}{3pt}       \setlength{\partopsep}{0pt}
			\setlength{\leftmargin}{3.5mm} \setlength{\labelwidth}{1em}
			\setlength{\labelsep}{0.5em} } }
	\newcommand{\squishend}{
\end{list}  }

\definecolor{codekw1}{rgb}{0,0.5,0}
\definecolor{codekw2}{rgb}{0.69,0,0.25}
\definecolor{codekw3}{rgb}{0.75,0,0}
\definecolor{codekw4}{rgb}{0.15, 0.5, 0.6}
\definecolor{at}{rgb}{0.64,0.29,0.64}
\definecolor{codeln}{rgb}{0.5,0.5,0.5}
\definecolor{codestr}{rgb}{0,0.6,0}

\lstdefinelanguage{myC}{
    morekeywords=[1]{interface},
    morekeywords=[2]{TypedTile, tTile},
    morekeywords=[3]{if, else, for, in, return, def, with, while},
    morekeywords=[4]{__global__, __device__, void, @overload, @property},
    morekeywords=[5]{size\_t, vector, set, Config, struct, TileShape, TileType, TensorExpr, class}, 
    morekeywords=[6]{get\_input, tTile-Operator},
    sensitive=true,
    morecomment=[l]{\#},
    morecomment=[l]{//},
    morecomment=[n]{/*}{*/},
    morestring=[b]",
    morestring=[b]',
    morestring=[b]"""
}

\begin{document}

\date{}
\title{\Large \bf \oursys: Fast Online Parallelism Switching for Elastic LLM Training}

\author{
	{\rm Yuanqing Wang$^{23}$\begin{NoHyper}\thanks{Equal contribution.}\end{NoHyper}, 
	     Yuchen Zhang$^{13}$\textsuperscript{\thefootnote},
	     Hao Lin$^{3}$,
	     Junhao Hu$^{3}$,
	     Chunyang Zhu$^{3}$,
		 Quanlu Zhang$^{3}$,
    }
    \vspace{-1mm}
	\and
	{\rm 
        Boxun Li$^{3}$,
        Guohao Dai$^{43}$,
        Zhi Yang$^{2}$,
	    Daning Cheng$^{1}$,
        Yunquan Zhang$^{1}$,
        Yu Wang$^{5}$,
    }
    \vspace{-1mm}
    \and
	{\rm 
         $^{1}$Institute of Computing Technology, CAS \hspace{0.3cm}
         $^{2}$Peking University \hspace{0.3cm}
         $^{3}$Infinigence AI \hspace{0.3cm}
    }
    \vspace{-1mm}
    \and
    {\rm
        $^{4}$Shanghai Jiao Tong University \hspace{0.3cm}
        $^{5}$Tsinghua University
   }
    \and 
    {\rm
    GitHub Repo: \url{https://github.com/infinigence/ElasticMegatron}
 }
}

\maketitle

\begin{abstract}
Modern large language model (LLM) training is inherently dynamic: resource fluctuations, RLHF phase shifts, and cluster elasticity continually reshape the optimal parallelism layout, posing a significant challenge to existing training frameworks built around a static execution model.
We present \oursys, a distributed training system for sub-second, online reconfiguration across arbitrary multi-dimensional parallelism.
At its core, we propose a Virtual Parameter Space (VPS) abstraction that unifies all distributed training states under one logical coordinate space, turning any parallelism configuration into a deterministic mapping and collapsing complex transition into manageable geometric intersections.
On top of VPS, a state routing-and-transition layer executes rank-local transfers under a memory-aware, deadlock-free schedule, and an Elastic Device Manager overlaps new-world construction with ongoing training to mask topology-change cost.
On dense and MoE models up to 235B parameters, \oursys reconfigures a 70B dense model in under 2\,s and a 235B MoE model in 4.36\,s, outperforming state-of-the-art checkpoint-based and elastic systems by up to three orders of magnitude while preserving correctness.
\end{abstract}


\section{Introduction}

Training large language models (LLMs) has become one of the most resource-intensive workloads on modern GPU clusters, 
    with state-of-the-art systems training billions and even trillions of parameters across tens of thousands of accelerators~\cite{megatron,deepspeed}. 
Sustaining this scale relies on a carefully composed stack of parallelism strategies, namely data, tensor, pipeline, expert, and optimizer-sharding parallelism~\cite{megatron,deepspeed,zero,fsdp}, which jointly partition parameters, gradients, and optimizer states to fit memory and achieve high throughput. 

This careful composition, however, rests on an implicit assumption: 
    that the parallelism layout chosen at startup remains optimal for the lifetime of the job. 
Mainstream training frameworks are built around this static execution model, 
    freezing cluster size, communicator groups, and memory layouts against a single configuration. 
In practice, this assumption is increasingly difficult to uphold. 
Production GPU clusters routinely reshape the pool of resources allocated to a single job, 
    as higher-priority workloads preempt resources, 
    spot capacity is revoked, 
    or co-located serving workloads release GPUs off-peak~\cite{gandiva,antman,pollux,varuna,bamboo,wagenlander2024tenplex}. 
Emerging workflows such as RLHF further require explicit parallelism transitions at runtime, 
    as GPUs are progressively shifted from rollout engines to training engines 
    once rollout demand subsides~\cite{rlhfuse,areal}. 
Across these scenarios, 
    the optimal parallel layout changes during training, 
    and a static system is forced to either leave GPUs idle or discard substantial progress to adapt.

The state-of-the-practice solution is to tear the job down and restart it under a new configuration. 
This \emph{checkpoint-and-restart} workflow serializes terabytes of model and optimizer states to a distributed file system, 
    reshapes and reloads them into a fresh training instance. 
Even with asynchronous checkpointing~\cite{megascale-nsdi24} 
    and parallelism-agnostic checkpoint formats~\cite{bytecheckpoint}, 
    loading a 70B model takes tens of seconds, 
    and re-initializing a multi-node communicator adds tens more. 
Existing elastic training systems~\cite{bamboo,oobleck-sosp23,wagenlander2024tenplex} 
    reduce this cost in specialized settings 
    but commit to pre-defined templates or a restricted subset of parallelism dimensions,
    falling short of the on-the-fly reconfiguration modern training demands.

As a more promising alternative, \emph{hot switching} redistributes model and optimizer states in place through the high-speed interconnects rather than storage, and without exiting the training job. 
However, generalizing hot switching for arbitrary parallelism transitions remains an unsolved system problem due to the following challenges.
First, modern frameworks shard each logical tensor simultaneously across DP, TP, PP, EP, and ZeRO dimensions, 
    producing heterogeneous state fragments 
    whose physical layout rarely aligns with tensor boundaries or across strategies. 
Second, transitioning between two multi-dimensional layouts is an $M$-to-$N$ peer-to-peer communication problem 
    that must be derived, ordered, and scheduled in a deadlock-free manner.
Third, execution is tightly constrained by GPU memory 
    as old and new states must transiently coexist with communication buffers.

In this paper, we present \oursys, 
    a distributed training system that enables online, sub-second reconfiguration of LLM training across arbitrary parallelism strategies. 
At the heart of \oursys is a new abstraction called the \emph{Virtual Parameter Space} (VPS), 
    a unified logical coordinate space in which every parameter, gradient, and optimizer state 
    is represented in its complete, unsharded form, 
    annotated with position and partitioning metadata. 
Under this abstraction, 
    every parallelism configuration becomes a deterministic mapping function 
    that projects the global VPS into a rank-local sub-VPS, 
    turning the heterogeneous physical states of different ranks into views of the \emph{same} logical object. 
Reconfiguring between strategies then reduces to geometric intersections 
    between source and destination sub-VPS regions, 
    avoiding the explosion of pairwise conversion rules.

On top of VPS, \oursys realizes hot switching through three cooperating components. 
A \emph{State Routing Planner} operates entirely in VPS coordinates 
    and derives, for every rank, the exact send, receive, and retain sets 
    required to move parameters and optimizer shards. 
A \emph{State Transition Engine} refines the planner's high-level plan into an optimized execution pipeline, 
    promoting point-to-point transfers into collective primitives where possible, 
    staging communication through memory-aware contiguous buffers, 
    and enforcing a provably deadlock-free schedule. 
An \emph{Elastic Device Manager} decouples logical strategies from physical communication groups,
    overlapping new-world initialization with ongoing training to mask the dominant cost of topology changes.

We implement \oursys as a pluggable middleware layer between PyTorch 
    and standard training frameworks such as Megatron-LM~\cite{megatron}. 
We evaluate it on a large cluster using both dense and MoE models ranging from 1.3B to 235B parameters.
Across DP, TP, PP, and EP transitions, 
    \oursys reconfigures 70B dense model in under 2\,s and 235B MoE model in 4.36\,s, 
    achieving up to 143$\times$ speedup over Megatron-LM distributed checkpointing for dense models, 
    up to 926$\times$ for MoE models, 
    and up to 15$\times$ over Tenplex~\cite{wagenlander2024tenplex} for cross-cluster task migration.

This paper makes the following contributions:
\squishlist
\item The Virtual Parameter Space (VPS) abstraction, 
    which unifies heterogeneous distributed training states under a single logical coordinate space 
    and reduces parallelism transitions to geometric intersection problems over VPS regions.
\item A cohesive set of mechanisms that realize efficient hot switching on top of VPS, 
    including State Routing Planner, 
    a memory-aware State Transition Engine with deadlock-free scheduling, 
    and an Elastic Device Manager that overlaps process-group reconfiguration with training.
\item \oursys, an end-to-end distributed training system that integrates these mechanisms and delivers sub-second online reconfiguration 
    with orders-of-magnitude speedups over prior approaches.
\squishend

\section{Background and Motivation}

Frontier LLM training is fundamentally a distributed systems problem, 
    built on a composition of data, tensor, pipeline, expert, and optimizer-sharding strategies across large GPU clusters. 
In practice, however, the best parallelism configuration is not fixed throughout training: 
    cluster capacity fluctuates and multi-phase workflows such as reinforcement learning (RL) shift resource demand over time. 
This section reviews these strategies and explains why fixing them for the lifetime of a job is increasingly untenable.

\subsection{Distributed LLM Training}

Because modern LLMs far exceed the capacity of a single GPU or server, 
    training systems must partition both model states and computation across many devices. 
Modern frameworks therefore combine several complementary parallelism strategies, 
    each slicing the workload along a different dimension 
    and relying on dedicated communication primitives to keep execution consistent.

\para{Data Parallelism (DP).} 
DP scales throughput by partitioning the global batch across devices while each device executes the same computation on a different data shard. 
Its memory footprint, however, remains bounded by what a single device can hold. 
Modern frameworks therefore adopt sharded data parallelism, such as ZeRO~\cite{zero} or FSDP~\cite{fsdp}, 
    which preserve the logical semantics of DP while partitioning parameters, gradients, and optimizer states across the DP group,
    materializing shards only when computation requires them.

\para{Model Parallelism (MP).} 
MP scales beyond a single-device model replica by partitioning the model itself across devices. 
Its two dominant forms are Tensor Parallelism (TP) and Pipeline Parallelism (PP). 
TP shards individual layers, reducing per-device memory at the cost of frequent intra-layer communication. 
PP instead partitions the model along depth, 
    assigning different groups of layers to different devices or stages. 
It uses micro-batches to overlap execution across stages 
    and substantially lowers the per-device memory footprint. 
Although PP introduces pipeline bubbles and scheduling complexity, 
    it remains a central technique for training very deep models at scale.

\para{Expert Parallelism (EP).} 
EP is specific to Mixture-of-Experts (MoE) models, where dense feed-forward layers are replaced by specialized experts. 
Unlike DP, TP, or PP, EP introduces an additional parallel dimension largely orthogonal to the rest of the model layout, 
    allowing expert parameters to be distributed independently from the dense components of the network. 
This decoupled structure makes parameter placement, communication, and runtime resharding substantially more irregular than in dense models.

By composing these strategies, 
    frameworks like Megatron-LM~\cite{megatron} and DeepSpeed~\cite{deepspeed} 
    can map massive workloads onto clusters with thousands of GPUs. 
Yet this scalability still rests on a static execution model: one parallelism configuration is chosen at startup, 
    and communicator groups, computation graphs, and memory layouts are built against that fixed world size.

\subsection{Elasticity of Training is Key to Efficiency}

The inability to adjust parallelism strategies at runtime creates a critical mismatch between static training systems and modern workloads. 
We highlight two key scenarios where dynamic parallelism switching is indispensable.

\para{Fluctuations in Resource Availability.} 
Training jobs in production GPU clusters rarely run against a fixed pool of resources~\cite{gandiva,antman,pollux}. 
Available GPU capacity expands and contracts as schedulers rebalance jobs, 
    higher-priority workloads preempt resources, or spot instances are revoked~\cite{antman,pollux,varuna,bamboo,wagenlander2024tenplex}. 
Co-location with LLM serving further amplifies this effect: peak traffic reserves large GPU allocations for inference, 
    while off-peak periods release substantial capacity back to the cluster. 
Static training frameworks cannot exploit these transient resources without costly checkpoint-and-restart overheads, 
    leading to poor cluster utilization~\cite{varuna,bamboo,wagenlander2024tenplex}. 
Elastic scaling lets a training job absorb these fluctuations and use transient capacity more effectively.


\para{Dynamic Workload Shifts in RLHF.} 
RLHF (Reinforcement Learning from Human Feedback) introduces a different kind of variability: 
    the demand of rollout and learning phases shifts over time. 
Modern RLHF typically decouples these phases, using dedicated inference engines (e.g., vLLM~\cite{vllm}) for rollout 
    and training engines (e.g., Megatron-LM~\cite{megatron}) for parameter updates. 
During rollout, substantial GPU capacity may initially be needed for prefill and decoding, 
    but this demand shrinks as sequences reach their End-of-Sequence tokens. 
Under a static allocation, rollout GPUs remain underutilized until the phase ends. 
To improve efficiency, these GPUs should be progressively released from the inference engine and reassigned to the learning engine, 
    requiring the training engine to expand its parallelism online.

\begin{figure}[t]
    \centering
    \includegraphics[width=0.65\linewidth]{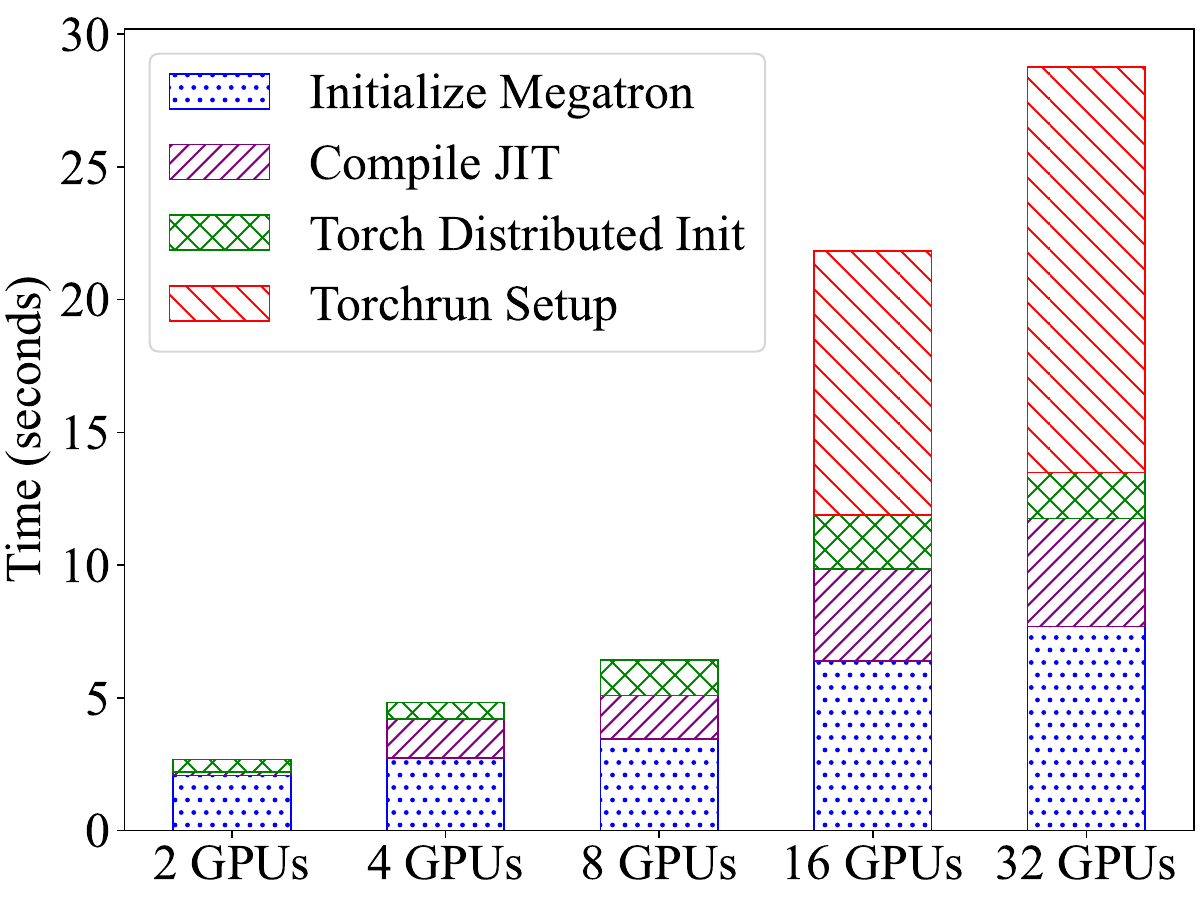}
    \caption{Init cost of Megatron-LM on different cluster scales.}
    \label{fig:init_cost}
\end{figure}

\subsection{Checkpoint-Restart is Costly}

The most straightforward way to support elasticity is an offline checkpoint-and-restart mechanism. 
The running job is halted, its model and optimizer states are written to a Distributed File System (DFS), 
    and a new training instance is launched to reload and reshape them under the target configuration. 
Although conceptually simple, this stop-and-go workflow is too expensive for real-time reconfiguration for three reasons:

\para{Storage I/O Bottlenecks.} The conventional approach to reconfiguration 
    relies on a \textit{Save-Transform-Load} cycle. 
This involves serializing terabytes of model parameters to a DFS, 
    utilizing offline CPU scripts to reshape tensors, 
    and reloading them into a new topology. 
This workflow is fundamentally bound by storage bandwidth. 
A single checkpoint operation can consume several minutes 
    (e.g., 2 minutes for slicing a LLaMA2-70B model from TP=8 to TP=4),
    which further scales linearly with the model size.

Asynchronous checkpointing hides part of the save latency~\cite{megatron,megascale-nsdi24}, 
    but storage remains on the reconfiguration path because states must still be reloaded and reshaped. 
Recent advancements like ByteCheckpoint~\cite{bytecheckpoint} 
    introduce parallelism-agnostic storage formats to enable load-time resharding. 
Yet, they remain constrained by the physical I/O throughput of the DFS, 
    rendering sub-second strategy switching impractical.  
For example, loading a LLaMA2-70B model costs about 30 seconds.


\para{Initialization Costs.} Even if state restoration were instantaneous, 
    cold-start initialization would still be expensive. 
Re-launching a distributed training job triggers a cascade of synchronous setup operations. 
\autoref{fig:init_cost} breaks down the initialization latency of Megatron-LM 
    across different cluster scales, excluding checkpoint loading. 
On a single node (up to 8 GPUs), 
    framework initialization dominates the overhead, taking up to 6 seconds. 
However, in multi-node deployments, 
    distributed setup becomes a severe bottleneck. 
Initializing a four-node cluster with 32 GPUs takes near 30 seconds, 
    heavily dominated by distributed initialization 
    (e.g., peer discovery, TCP store synchronization, and communication group construction). 
These startup costs make frequent, fine-grained strategy changes prohibitive, 
    largely negating the benefits of dynamic scaling for short-lived fluctuations.


\para{Rigidity of Current Elastic Approaches.} Recent elastic systems reduce some of this cost, 
    but rely heavily on pre-defined templates or fixed scaling schedules~\cite{bamboo, oobleck-sosp23, wagenlander2024tenplex}. 
They typically support only restricted subsets of parallelism and cannot handle arbitrary topology changes 
    or general multi-dimensional tensor redistribution. 
    Consequently, general-purpose dynamic scaling remains an unsolved challenge 
    in distributed training system design.

\subsection{Hot Switching of Training is Challenging}
\label{subsec:challenge}
To bypass the prohibitive overheads of checkpoint-restart, 
    a promising alternative is hot switching, 
    which redistributes model and optimizer states directly through the high-speed network without exiting the training job. 
However, enabling seamless in-memory transitions between arbitrary multi-dimensional strategies remains an unsolved problem due to three primary challenges.

\para{C1: Complex State Representation and Heterogeneity.}
As model sizes grow and architectures become more complex, 
    state management in distributed training frameworks becomes highly intricate. 
To achieve optimal performance and memory efficiency, 
    mainstream frameworks introduce advanced optimizations 
    that heavily complicate the physical state. 
For instance, in the naive implementation of DP, 
    each device holds a complete replica of the optimizer state, 
    making the DP instances largely independent and homogeneous. 
However, modern memory optimizations like ZeRO 
    partition the optimizer states and gradients across DP instances 
    to significantly reduce memory footprint. 
While highly efficient, 
    this breaks the homogeneity among devices, 
    leading to a unique, disjoint shard of the model state on each GPU. 
Consequently, accurately identifying and representing the correct initial and final states 
    during a topology switch requires tracking highly heterogeneous shards 
    across multiple dimensions, 
    making static state management extremely difficult.

\para{C2: Intricate Routing and Mapping.}
Establishing the connection between abstract logical states 
    and concrete physical locations poses another major challenge. 
Switching from an old parallelism configuration to a new one 
    is not a simple data broadcast; 
    it requires a precise mathematical mapping 
    between the source and destination physical shards. 
Because a single tensor might be sliced differently across TP, PP and ZeRO dimensions, 
    redistributing these parameters necessitates a highly complex 
    $M$-to-$N$ peer-to-peer communication pattern. 
These operations must be strictly ordered between interacting peers, 
    and any mismatch in execution sequence or payload size results in distributed hangs. 
This hazard is exacerbated 
    as individual ranks often lack a global view of the topology shift. 
Thus, deriving a correct, optimal, and deadlock-free routing strategy 
    to move these shards across the physical network topology 
    is a non-trivial orchestration problem.

\para{C3: Memory-Constrained Execution.}
Executing online state transitions is fundamentally constrained 
    by the rigid limits of GPU memory. 
For instance, the training states excluding activations for a single DP replica 
    of a 70B model under standard \texttt{bf16-fp32} mixed-precision training 
    consume approximately 1.26\,TB. 
During the transition phase, 
    if the system naively attempts to materialize both the old and new parameter states simultaneously 
    alongside large communication buffers, 
    it will easily triggers the Out-Of-Memory (OOM) failure. 
The execution engine must safely navigate these strict memory constraints 
    while minimizing the end-to-end switching latency. 
This guarantees that hardware training throughput remains entirely uncompromised.


Together, these challenges show that hot switching is not merely a matter of moving tensors between devices. 
It requires a unified state representation, a correct routing mechanism across changing layouts, 
    and an execution engine that respects tight memory limits throughout the transition. 
Any practical online switching system must satisfy all three simultaneously.

\section{\oursys Overview}
\label{sec:overview}

\oursys enables millisecond-level online reconfiguration of LLM parallelism strategies. 
To handle state heterogeneity across strategies, 
    \oursys introduces the \textit{Virtual Parameter Space} (VPS), 
    a unified logical-physical indexing abstraction over the full model. 
With VPS, parameter movement becomes a geometric transformation between parallel layouts, 
    enabling exact state tracking and efficient transition planning.


\begin{figure}[ht]
    \centering
    \includegraphics[width=0.8\linewidth]{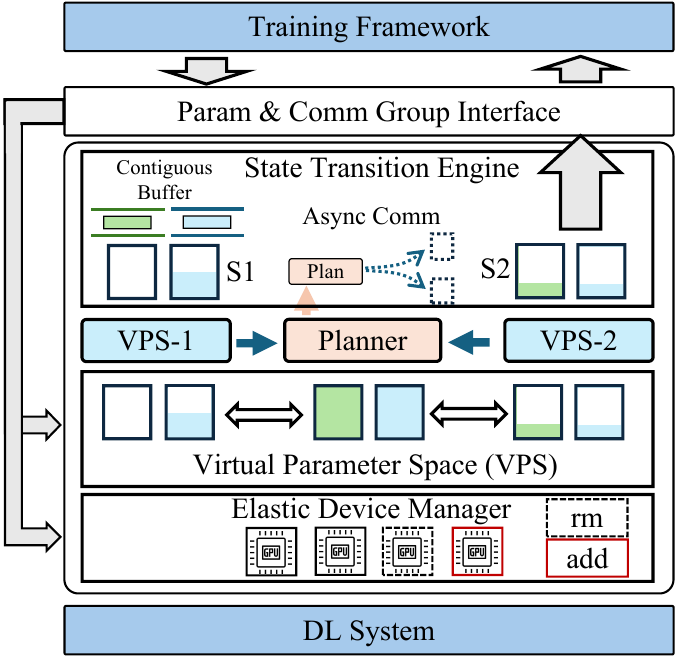}
    \caption{\oursys architecture overview.}
    \label{fig:system-arch}
\end{figure}

\autoref{fig:system-arch} shows the architecture of \oursys. 
\oursys operates as a pluggable middleware layer between the underlying deep learning system 
    (e.g., PyTorch) and the training framework (e.g., Megatron-LM~\cite{megatron}), 
    remaining transparent to the upper framework. 
A typical strategy transition flows through four core components.

\para{VPS.} The VPS tracks the meta-information of all parameters, 
    providing each rank with a holistic view of the complete model under any parallel strategy. 
During a transition from strategy \texttt{S1} to \texttt{S2}, 
    each rank instantiates a local sub-VPS 
    corresponding to the exact parameters distributed to it 
    under the respective strategies (\texttt{VPS-1} and \texttt{VPS-2}).

\para{State Routing Planner.} In \oursys, 
    switching parallelism strategies fundamentally equates to routing distributed training states 
    across changing parallel layouts. 
Within the VPS, the planner maps parameters, optimizer states, and other migratable state shards 
    from their source physical layout to their destination layout. 
It mathematically derives the required $M$-to-$N$ transmission matrices 
    by comparing the logical coordinate intersections 
    between the source (\texttt{VPS-1}) and destination (\texttt{VPS-2}) layouts.

\para{State Transition Engine.} Once the routing plan is generated, 
    it is dispatched to the transition engine for optimization and execution. 
The engine refines the naive routing plan into an efficient communication procedure 
    by optimizing communication primitives, 
    coalescing fragmented transfers, 
    and scheduling state movement under hardware constraints. 
Because hardware memory is strictly constrained during a transition, 
    the engine utilizes a phased, asynchronous scheduling mechanism. 
It carefully orchestrates the lifecycle of send/receive buffers, 
    maximizing communication overlap 
    while strictly bounding transient memory peaks to OOM.


\para{Elastic Device Manager.} Serving as the foundational hardware and communication substrate of \oursys, 
    the Elastic Device Manager (EDM) provides a unified abstraction over dynamic cluster resources. 
It manages the complete lifecycle of distributed communication groups 
    by orchestrating their background establishment, caching, and instantaneous hot switching 
    to seamlessly supply the training framework with the correct physical topology. 
Furthermore, the EDM handles cluster elasticity 
    including resource fluctuation 
    by dynamically managing node joins and leaves, 
    paving the way for seamless logical state migration.

Ultimately, these components work in concert 
    to decouple strategy formulation from physical execution, 
    providing a robust, transparent engine for elastic LLM training.




\section{System Design}
\label{sec:design}

This section presents the design of \oursys. 
We first introduce the Virtual Parameter Space (VPS) abstraction (\S\ref{sec:vps_abstraction}), 
    the unified representation of distributed training states. 
We then present the state routing planner (\S\ref{subsec:routing_planner}), 
    for deriving routing plans that migrate distributed training states across parallelism transitions, 
    and the state transition engine (\S\ref{subsec:execution_engine}), 
    for optimizing and executing these transitions efficiently under tight memory constraints. 
Finally, we describe the Elastic Device Manager (\S\ref{sec:elastic_device_management}), 
    responsible for managing communication topologies for elastic scaling.

\subsection{Virtual Parameter Space}
\label{sec:vps_abstraction}

To address the inherent heterogeneity of model states 
    across diverse parallelism strategies (C1 in \autoref{subsec:challenge}) 
    and to systematically resolve the intricate $M$-to-$N$ routing problem (C2), 
    we introduce the core abstraction of our system: 
    the \textbf{Virtual Parameter Space (VPS)}.

\begin{figure}[t]
    \centering
    \includegraphics[width=0.8\linewidth]{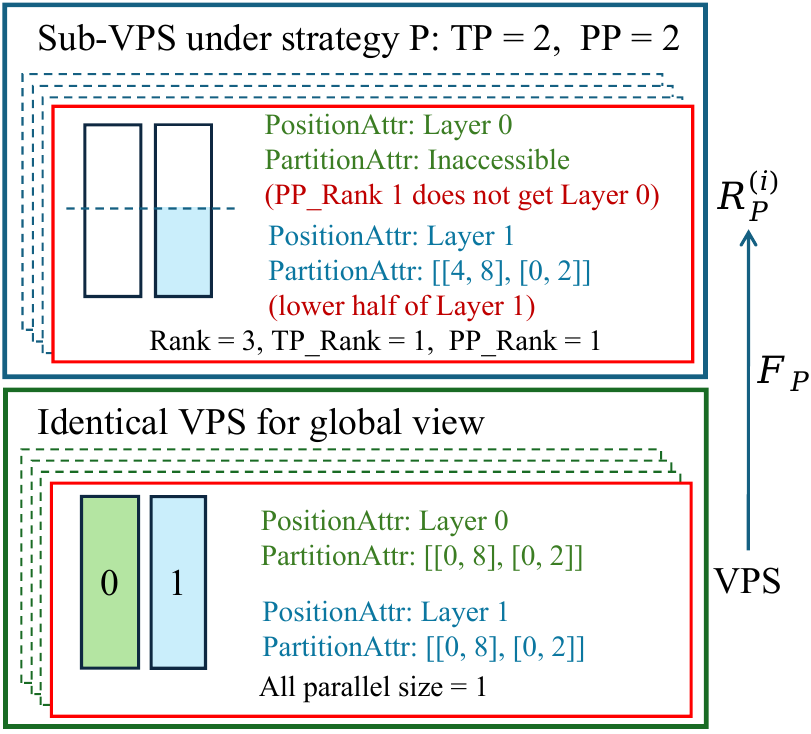}
    \caption{VPS mapping from a shared global view to a rank-local sub-VPS. The bottom half shows the identical global VPS seen by all ranks, with each parameter represented in its complete logical form. The top half shows how a parallel strategy $\mathcal{P}$ projects this space into the sub-VPS of one rank.}
    \label{fig:vps_mapping}
\end{figure}

The key observation behind VPS is that, regardless of sharding, 
    parameters, gradients, and optimizer states still belong to the same underlying model. 
VPS therefore replaces heuristic pairwise conversion rules between strategies 
    with a single strategy-agnostic intermediate representation.

\para{Definition of VPS.}
We define the VPS as a unified, logical, and contiguous coordinate space 
    where sizes of all parallel dimensions are intrinsically set to 1, 
    as depicted at the bottom of ~\autoref{fig:vps_mapping}. 
In this space, 
    every tensor (e.g., a weight matrix or an optimizer variance vector) 
    exists in its complete, unsharded form, 
    represented by a logical multi-dimensional index bounding box. 

For any given parallelism configuration $\mathcal{P}$ 
    (incorporating DP, TP, PP, EP, and ZeRO parallelism degrees), 
    the configuration acts as a mathematical mapping function $F_{\mathcal{P}}$. 
This function maps a specific rank residing on hardware device $d_i$ 
    to a distinct coordinate range (a sub-tensor bounding box) within the VPS. 
Formally, the physical state on device $d_i$ under strategy $\mathcal{P}$ 
    corresponds to a logical VPS region $R_{\mathcal{P}}^{(i)}$:
    $R_{\mathcal{P}}^{(i)} = F_{\mathcal{P}} \left ( \text{VPS},\ d_{i}\right ) \subset \text{VPS}$.

As shown in the bottom half of \autoref{fig:vps_mapping}, 
    every rank starts from the same global VPS, 
    where parameters are represented in their complete logical form 
    and all parallel dimensions are collapsed to size 1. 
Each parameter carries \uline{position attributes}, which identify its semantic role in the model, 
    and \uline{partitioning attributes}, which describe how it is physically distributed under a given strategy. 

In the top half of \autoref{fig:vps_mapping}, 
    this global space is projected through $F_{\mathcal{P}}$ into the \emph{sub-VPS} of a specific rank. 
The sub-VPS preserves the full model metadata but materializes only the regions assigned to that rank; 
    parameters outside the rank's PP stage or other partitions remain visible but are marked inaccessible. 
For accessible parameters, the sub-VPS records the exact local partition attributes of the resident shard, 
    enabling uniform reasoning about both absent and resident states during a transition.

\begin{figure}[htbp]
    \centering
    \includegraphics[width=0.8\linewidth]{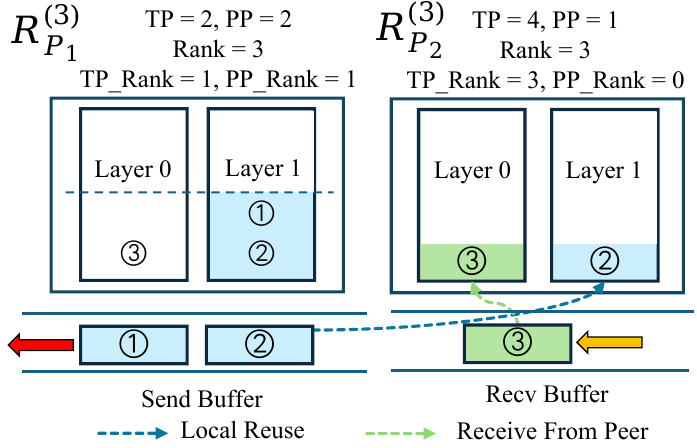}
    \caption{State routing plan derived from VPS for rank 3 switching from $(\mathrm{TP}, \mathrm{PP}) = (2, 2)$ to $(4, 1)$. The planner identifies slices to send (\ding{172}), retain locally (\ding{173}), and receive (\ding{174}).}
    \vspace{-15pt}
    \label{fig:comm_plan}
\end{figure}

\subsection{State Routing Planner}
\label{subsec:routing_planner}

By abstracting physical shards into logical regions within the VPS, 
    the complex challenge of routing distributed training states 
    across a parallelism transition is effectively reduced 
    to a geometrical intersection problem.

\subsubsection{Plan for Parameters}

Consider a system transitioning from a source strategy $\mathcal{P}_{src}$ 
    to a destination strategy $\mathcal{P}_{dst}$. 
A device $i$ holds logical regions $R_{src}^{(i)}$ and $R_{dst}^{(i)}$ 
    within the VPS under the source and destination configurations, respectively. 
To determine the exact data payload that must be transmitted, 
    our state routing planner computes the mathematical union of these logical regions:
    $T_{\mathcal{P}_{src} \rightarrow \mathcal{P}_{dst}}^{(i)} = R_{src}^{(i)} \cup R_{dst}^{(i)}$.

\autoref{fig:comm_plan} illustrates rank 3 switching from $(\text{TP}, \text{PP}) = (2, 2)$ to $(4, 1)$. 
Comparing $R_{src}^{(3)}$ and $R_{dst}^{(3)}$ reduces the transition to three slice types in 
    $T_{\mathcal{P}_{src} \rightarrow \mathcal{P}_{dst}}^{(3)}$: 
    slices to send (\ding{172}), retain locally (\ding{173}), and receive (\ding{174}).

To transition to $\mathcal{P}_{dst}$, 
    device $i$ must receive parameters absent from $R_{src}^{(i)}$ 
    and send specific parameters to other devices $j$ 
    that require them for $R_{dst}^{(j)}$. 
Since some parameter slices may be retained locally across the reconfiguration, 
    we partition $T_{\mathcal{P}_{src} \rightarrow \mathcal{P}_{dst}}^{(i)}$ 
    into three mutually exclusive subsets:

\begin{equation*}
    T_{\mathcal{P}_{src} \rightarrow \mathcal{P}_{dst}}^{(i)} = 
    \underbrace{\left ( R_{src}^{(i)} \setminus R_{dst}^{(i)} \right )}_{T_{\texttt{send}}^{(i)}} 
    \bigcup \underbrace{\left ( R_{dst}^{(i)} \setminus R_{src}^{(i)} \right )}_{T_{\texttt{recv}}^{(i)}}
    \bigcup \underbrace{\left ( R_{src}^{(i)} \cap R_{dst}^{(i)} \right )}_{T_{\texttt{retain}}^{(i)}}
\end{equation*}

Here, $T_{\texttt{send}}^{(i)}$, $T_{\texttt{recv}}^{(i)}$, and $T_{\texttt{retain}}^{(i)}$ 
    represent the parameters device $i$ must send to other peers, 
    receive from other peers, 
    and retain locally, respectively. 
In \autoref{fig:comm_plan}, 
    they correspond to \ding{172}, \ding{174}, and \ding{173}. 
For every parameter slice requiring transmission, 
    the source and destination devices are deterministically resolved via the VPS, 
    as it inherently encodes the global distribution mappings for any configuration. 
For data parallelism, 
    where a tensor might have redundant sources or destinations, 
    \oursys applies a proximity-based selection policy (discussed in \S\ref{subsec:execution_engine}). 
Ultimately, the routing planner maps these logical categories 
    into concrete local actions on each rank's buffers,
    and then automatically generates precise point-to-point communication pairs 
    (\texttt{Send/Recv}) for exact data slices, 
    establishing a functionally correct baseline state routing plan.

\begin{figure}[h]
    \centering
    \includegraphics[width=0.7\linewidth]{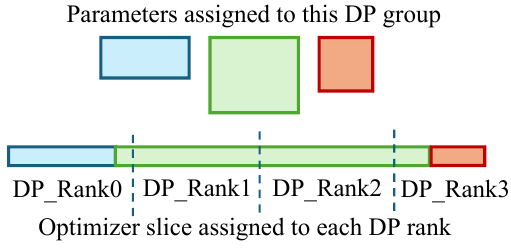}
    \caption{Unaligned optimizer states in a ZeRO DP group.}
    \label{fig:zero_opt_slice}
    \vspace{-15pt}
\end{figure}

\subsubsection{Plan for Optimizer}

Optimizer state requires special handling during strategy switching, 
    especially for stateful optimizers and mixed-precision training where optimizer states and master weights persist in GPU memory.

Traditionally, scaling data parallelism proportionally scales optimizer state replication. 
ZeRO~\cite{zero} mitigates this by sharding optimizer states across the DP group 
    (i.e., ranks processing the same model partition across different data batches), 
    significantly reducing memory footprint without precision degradation. 
As depicted in \autoref{fig:zero_opt_slice} 
    (illustrating ZeRO inside Megatron-LM~\cite{megatron}), 
    all parameters assigned to a DP group are first flattened into a single contiguous buffer, 
    and the optimizer states are then evenly partitioned across DP ranks by contiguous buffer ranges. 
As a result, the optimizer slice owned by one DP rank is defined by flat offsets in that buffer, 
    rather than by semantic parameter boundaries. 
In the figure, 
    the colored parameter tensors at the top are logical model tensors, 
    whereas the slices at the bottom are the actual optimizer-state ranges assigned to DP ranks. 
Because these rank-local ranges can cut across tensor boundaries, 
    a single semantic parameter tensor may have its optimizer states scatter across multiple DP ranks. 
This misalignment between semantic tensors and physical shards 
    significantly complicates optimizer state migration during a strategy transition.

\begin{figure}[h]
    \centering
    \includegraphics[width=0.8\linewidth]{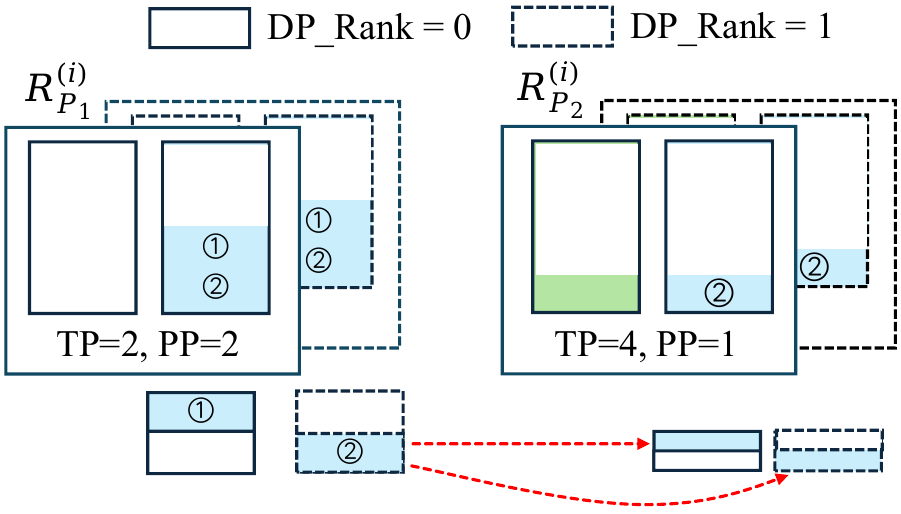}
    \caption{State routing plan for sharded optimizer states. The solid and dashed boxes denote \texttt{DP\_Rank}=0 and \texttt{DP\_Rank}=1 for the same TP/PP rank under the source and destination strategies. Only the transfer of slice \ding{173} is shown.}
    \label{fig:comm_plan_zero}
    \vspace{-15pt}
\end{figure}

\oursys leverages the VPS abstraction to resolve this mismatch systematically. 
The key step is to map each flat optimizer shard 
    back into the same logical VPS (which is also a contiguous, flat space) coordinate space used for parameters. 
Although the optimizer is physically partitioned by contiguous buffer offsets, 
    those offsets still correspond to a set of logical parameter regions in the flattened parameter order. 
Thus, for every source or destination \texttt{DP\_Rank}, 
    \oursys can invert the flat buffer assignment into a collection of VPS subregions 
    and reason about them exactly as it does for ordinary parameter shards. 
Once expressed in VPS coordinates, 
    boundary-mismatched optimizer slices become ordinary geometric intersections between source and destination regions, 
    rather than a special-case reshaping problem.

\autoref{fig:comm_plan_zero} illustrates this process 
    under the same strategy transition as \autoref{fig:comm_plan}. 
The solid and dashed outlines denote \texttt{DP\_Rank}=0 and \texttt{DP\_Rank}=1 
    for the same TP/PP location under the source and destination strategies. 
In the source configuration, 
    both DP ranks hold the same parameter shard, 
    but the corresponding optimizer state is split across them into slices \ding{172} and \ding{173}. 
The VPS enables each rank to determine which logical portion of its local optimizer slice 
    remains local in the destination layout and which portion must move to another \texttt{DP\_Rank}. 
As shown, 
    slice \ding{173} is routed from its source owner to the destination owner; 
    slice \ding{172} follows the same logic. 
This demonstrates how VPS bridges the gap between semantic parameter tensors and flat buffer boundaries.

\subsubsection{Plan for Dataset and Other States}

To preserve training semantics across strategy transitions, 
    dataset states must remain synchronized. 
When parallelism degrees change, 
    \oursys adjusts global batch sizes and micro-batch allocations 
    and tracks the number of samples consumed by each DP rank before the switch, 
    ensuring that the data loader resumes without skipping or duplicating samples.

Other training states, such as hyperparameters and scalar metrics, are small. 
\oursys treats rank 0 as the source of truth and broadcasts them to the new communication groups, 
    ensuring global consistency with negligible overhead.




\begin{algorithm}[!t]
\normalem
\SetInd{0.61em}{0.61em}
\caption{Communication Orchestration}
\label{alg:orchestration}
\small
    \definecolor{deepgreen}{rgb}{0.0, 0.4, 0.0}
    \SetKwComment{Comment}{$\triangleright$\ }{\ }
    \newcommand{\mycommfont}[1]{\footnotesize\ttfamily\textcolor{deepgreen}{#1}}
    \SetCommentSty{mycommfont}

    \SetKwInput{KwInput}{Input}
    \SetKwInput{KwOutput}{Output}
    \SetKwFunction{FSwitch}{ExecuteSwitch}
    \SetKwFunction{FChunk}{MemoryAwareChunk}
    \SetKwFunction{FOptimize}{OptimizePrimitives}
    \SetKwProg{Fn}{Function}{:}{}

\KwInput{Plan $\Pi_{\text{naive}}$, memory $M_{\text{avail}}$, rank $i$, total ranks $N$}
\KwOutput{Instantiated destination states $S_{\text{dst}}^{(i)}$}

\Fn{\FSwitch{$\Pi_{\text{naive}}$, $M_{\text{avail}}$, $i$, $N$, $G$}}{
    $\Pi_{\text{opt}} \gets$ \FOptimize{$\Pi_{\text{naive}}, G$}\;
    \text{FreeObsoleteBuffers}$()$\;
    $Steps \gets \{1, 2, \dots, N-1\}$\;
    $Stages \gets$ \FChunk{$\Pi_{\text{opt}}, Steps, M_{\text{avail}}$}\;

    \ForEach{$stage \in Stages$}{
        $B_{\text{send}}, B_{\text{recv}} \gets \text{AllocateContinuousBuffers}(stage)$\;
        $\text{PackData}(B_{\text{send}}, stage)$\;
        
        \ForEach{$s \in stage$}{
            \Comment{XOR-based deadlock-free scheduling}
            $peer \gets i \veebar s$\; 

            \If{$peer < N$}{
                $\text{AsyncSend}(B_{\text{send}}[peer], peer)$\;
                $\text{AsyncRecv}(B_{\text{recv}}[peer], peer)$\;
            }
        }
        
        $\text{SynchronizeAll}()$\;
        $\text{UnpackData}(B_{\text{recv}}, S_{\text{dst}}^{(i)})$\;
        $\text{Free}(B_{\text{send}}, B_{\text{recv}})$\;
    }

    \Return{$S_{\text{dst}}^{(i)}$}\;
}

\Fn{\FChunk{$\Pi_{\text{opt}}$, $Steps$, $M_{\text{avail}}$}}{
    $M_{\text{global\_min}} \gets \text{AllReduce}(M_{\text{avail}}, \text{op}=\text{MIN})$\;
    $Stages, \ current\_stage \gets \varnothing$\;
    $current\_mem \gets 0$\;
    
    \ForEach{$s \in Steps$}{
        $cost_s \gets \text{CalculateMemCost}(\Pi_{\text{opt}}, s)$\;
        \eIf{$current\_mem + cost_s > M_{\text{global\_min}}$}{
            $Stages \gets Stages \cup \{current\_stage\}$\;
            $current\_stage \gets \{s\}$\;
            $current\_mem \gets cost_s$\;
        }{
            $current\_stage \gets current\_stage \cup \{s\}$\;
            $current\_mem \gets current\_mem + cost_s$\;
        }
    }
    \If{$current\_stage \neq \varnothing$}{
        $Stages \gets Stages \cup \{current\_stage\}$\;
    }
    
    \Return{$Stages$}\;
}

\Fn{\FOptimize{$\Pi_{\text{naive}}$, $G$}}{
    $\Pi_{\text{pruned}}, \Pi_{\text{opt}} \gets \text{TopologyAwarePrune}(\Pi_{\text{naive}}, G), \varnothing$\;
    $\mathcal{T} \gets \text{GroupByLogicalTensor}(\Pi_{\text{pruned}})$\;
    
    \ForEach{$t \in \mathcal{T}$}{
        $srcs, dsts, slices \gets \text{UniqueSrcDstRange}(t)$\;
        $cont = \text{IsContiguous}(slices)$\;
        \uIf{$|srcs| == 1$ \textbf{and} $|dsts| > 1$ \textbf{and} \text{IsIdentical}$(slices)$}{
            $\Pi_{\text{opt}} \gets \Pi_{\text{opt}} \cup \{\text{Broadcast}(t)\}$\;
        }
        \uElseIf{$|srcs| == 1$ \textbf{and} $|dsts| > 1$ \textbf{and} $cont$}{
            $\Pi_{\text{opt}} \gets \Pi_{\text{opt}} \cup \{\text{Scatter}(t)\}$\;
        }
        \uElseIf{$|srcs| > 1$ \textbf{and} $|dsts| == 1$ \textbf{and} $cont$}{
            $\Pi_{\text{opt}} \gets \Pi_{\text{opt}} \cup \{\text{Gather}(t)\}$\;
        }
        \Else{
            $\Pi_{\text{opt}} \gets \Pi_{\text{opt}} \cup t$\;
        }
    }
    \Return{$\Pi_{\text{opt}}$}\;
}
\end{algorithm}
%

\subsection{State Transition Engine}
\label{subsec:execution_engine}

The state routing planner produces a high-level transition plan. 
The State Transition Engine refines and executes this plan under strict memory and communication constraints. 
As shown in \autoref{alg:orchestration}, 
    it optimizes communication primitives, stages transfers through dedicated buffers, 
    and schedules execution to bound both latency and transient memory overhead.

\para{Communication Primitive Optimization.} While the VPS provides a correct baseline point-to-point communication plan ($\Pi_{\text{naive}}$), 
    this initial plan is agnostic to the underlying network topology. 
The engine optimizes this plan prior to execution (\autoref{alg:orchestration}, Line 2). 
First, when altering data parallelism degrees, 
    a required parameter in $R_{dst}^{(i)}$ may have multiple valid sources $R_{src}^{(j)}$ across different devices. 
\oursys automatically selects the source-destination pair 
    that maximizes intra-node topology bandwidth 
    (e.g., favoring NVLink over PCIe or inter-node Ethernet).

Second, the system promotes the remaining point-to-point operations 
    into collective primitives such as \texttt{scatter}, \texttt{gather}, and \texttt{broadcast} (Line 37-47). 
Because the VPS exposes both the logical ranges and physical mappings of each transfer, 
    \textsc{\oursys} can infer collective intent deterministically by grouping operations by logical tensor 
    and analyzing source/destination cardinalities together with slice contiguity.
By mathematically deriving these topological patterns from the VPS, 
    \textsc{\oursys} minimizes instruction queue depths 
    and maximally exploits hardware-accelerated collective libraries (e.g., NCCL) 
    without requiring manual developer annotations.

\para{Memory-Aware Contiguous Buffer.} 
During resharding, source and destination states must transiently coexist, 
    severely constraining GPU memory. 
The engine mitigates this pressure through a memory-aware staging pipeline. 
Before transitioning, it garbage-collects obsolete buffers such as unused gradients and activations (Line 3). 
Using VPS to estimate payload sizes and peak memory demand, 
    the engine performs a pre-flight memory check (Line 22) and, if needed, 
    partitions the logical transmission steps ($s$) into finer-grained stages. 

Within each stage, 
    the raw communication plan often yields fragmented point-to-point transfers, 
    which degrade network bandwidth and incur severe kernel launch and synchronization overheads. 
To address this, 
    \oursys allocates contiguous, dedicated send/receive buffers per communicating peer (Line 7) 
    and packs the scheduled slices prior to transmission (Line 8). 
Because intra-GPU memory bandwidth vastly exceeds interconnect throughput, 
    the performance gained from coalesced transfers far outweighs the packing overhead. 
Finally, \oursys eagerly deallocates source parameters 
    the moment a stage's transfer completes and the data is unpacked (Line 18). 
This precise lifecycle management strictly bounds peak memory consumption, 
    effectively eliminating OOM failures during online strategy switching.

\para{Asynchronous Deadlock-Free Communication.}
With data coalesced into dedicated buffers, 
    the engine issues asynchronous batch communication to overlap transfers across peers (Lines 9-15). 
To prevent deadlocks, it orchestrates communication through a logical step-based schedule. 
Let $N$ denote the total number of ranks participating in the transition. 
The schedule proceeds over logical steps $s \in \{1, 2, \dots, N-1\}$. 
In each step, rank $i$ communicates with at most one peer, 
    whose rank index is determined by the XOR rule: $\text{Peer}\left (i, s\right ) = i \veebar s = j$.
If $j \geq N$ or $j$ is not in the communication plan of $i$ ($j \notin R_{src}^{(i)} \cup R_{dst}^{(i)}$), 
    the pair is invalid and no communication is issued in that step. 
Otherwise, due to the symmetry of the XOR operation and the deterministic nature of the schedule, 
    the target rank $j$ concurrently resolves back to rank $i$: $\text{Peer}\left (j, s\right ) = j \veebar s = (i \veebar s) \veebar s = i$.
By mathematically pairing communicating peers, 
    separating send and receive buffers, 
    and enforcing this async schedule, 
    \oursys guarantees a deadlock-free and high-throughput communication sequence.

\subsection{Elastic Device Manager}
\label{sec:elastic_device_management}
To support fluid transitions between arbitrary parallelism strategies 
    and adapt to cluster-level elasticity, 
    \oursys introduces the Elastic Device Manager (EDM). 
The EDM decouples logical strategy formulation from physical communication-state management, 
    providing unified orchestration for both strategy switching and device scaling.

\para{Lifecycle Management of Communication Groups.} Communication groups are essential to distributed training. 
Traditional engines initialize a rigid global group (\texttt{World}) 
    and carve out fixed sub-groups tied to a static strategy. 
\oursys breaks this rigidity via the EDM, 
    which encapsulates the communication groups associated with each strategy 
    and binds them to its VPS. 
During a strategy transition, 
    the EDM orchestrates an in-place update mechanism: 
    it automatically instantiates or retrieves cached communication groups required by the new strategy, 
    updates process ranks across parallel dimensions, 
    and re-binds communication buffers prior to data transmission. 
This decoupled state management ensures that switching communication topologies 
    remains completely transparent to the upper-level routing pipelines.

\para{Dynamic Scaling and Overlap.} Depending on cluster dynamics, 
    the EDM supports two reconfiguration modes:

\textit{In-Place Reconfiguration.} When the global set of physical processes remains static 
    but workload characteristics fluctuate, 
    the EDM reconfigures within the current device pool. 
This is ideal for workflows like RLHF, 
    where the compute ratio between generation and training phases shifts dynamically, 
    or when processing datasets with high sequence length variance. 
The EDM instantly swaps active process groups from its cache, 
    tailoring the parallel strategy on the fly 
    to maintain high throughput without reallocating physical resources.

\textit{Elastic Device Scaling with Overlapping Initialization.} 
When responding to predictable resource fluctuations 
    (e.g., premeditated elasticity or spot instance scheduling), 
    the cluster topology inherently changes as nodes join or leave. 
Establishing a new \texttt{World} process group 
    typically incurs significant synchronous overhead 
    (often dominating the latency shown in \autoref{fig:init_cost}). 
To minimize this end-to-end switching latency, 
    the EDM employs an asynchronous overlapping mechanism. 
It permits the training loop to continue uninterrupted on the existing communication topology 
    while the new \texttt{World} and its required sub-groups are initialized in the background. 
Once the new infrastructure is fully established, 
    the EDM synchronizes the global state, migrates the necessary tensors, and resumes training, 
    drastically reducing the wall-clock penalty of elastic scaling.

\vspace{-5pt}
\section{Implementation}
\vspace{-5pt}
\label{sec:implementation}
\oursys is implemented in 7K lines of Python code as a pluggable middleware layer for Megatron-LM.
The integration boundary is intentionally narrow to reduce intrusiveness to training frameworks: Megatron registers with \oursys the training-state tensors it wants managed during reconfiguration, together with their state types (e.g., parameters, gradients, and optimizer states) and the metadata needed to map them into VPS and track their distributed layouts.
In return, \oursys exposes runtime interfaces for constructing, caching, and acquiring the communicator groups required by different parallel strategies, allowing Megatron to invoke reconfiguration without rewriting its core training logic.

\para{Releasing Contiguous Buffers with Distributed References.} Modern LLM training engines allocate large contiguous buffers and slice them into individual tensors to minimize fragmentation and optimize communication during gradient, optimizer, and weight updates. However, this complicates manual memory management: buffers cannot be garbage-collected until all distributed references across the codebase are deleted, making centralized tracking intractable. \oursys addresses this by explicitly deallocating the underlying physical memory at a low level immediately after transmission, leaving high-level references intact. By strictly enforcing transmission and access orderings, \oursys efficiently reuses memory without requiring backend engine modifications or causing illegal memory accesses.

\section{Evaluation}

We evaluate \oursys in two representative system scenarios: \emph{parallelism reconfiguration}, where a running job switches among different parallelism layouts on the current device pool, and \emph{elastic resource scaling}, where the set of allocated devices changes and training must adapt accordingly. 
We further validate correctness by examining whether repeated parallelism switching affects training convergence. We conclude with an ablation study evaluating the performance gains from tensor buffering and asynchronous communication within our EDM.

\subsection{Experimental Setup}

\para{Testbed.} We conduct our end-to-end evaluations on a 16-node GPU cluster. Each node is equipped with $8\times$ NVIDIA A800 (80\,GB) GPUs, dual Intel Xeon Platinum 8358 CPUs (2.6\,GHz, 64 cores), and 1\,TB of main memory. Intra-node GPU communication is facilitated by NVLink (400\,GB/s). For inter-node communication, each node is provisioned with eight Mellanox ConnectX-5 RDMA NICs, providing 200\,Gbps of network bandwidth per interface.


\para{Baselines.} We compare \oursys against three representative systems for elastic training and reconfiguration. Tenplex~\cite{wagenlander2024tenplex} supports elastic training through checkpoint-based task migration and reconfiguration across parallel layouts. HotSpa~\cite{hotspa} is an in-memory hot-switching system that relies on pre-registered strategies and static execution graphs for parameter switching. Megatron-LM Distributed Checkpointing (MCP)~\cite{megatron} is a checkpoint-based baseline natively supported by Megatron-LM.

\para{Models and Datasets.} We evaluate \oursys using three representative LLM families. For dense architectures, we use GPT-3 variants (1.3B, 2.7B, and 6.7B parameters) and LLaMA-2 variants (7B, 13B, and 70B). For MoE architectures, we use the widely adopted Qwen3-MoE (30B-A3B and 235B-A30B). For convergence experiments, all models are trained on the English Wikipedia corpus~\cite{wiki-cite}, a standard dataset for large-scale language model pre-training.


\begin{figure}[H]
    \centering
    \includegraphics[width=1.0\linewidth]{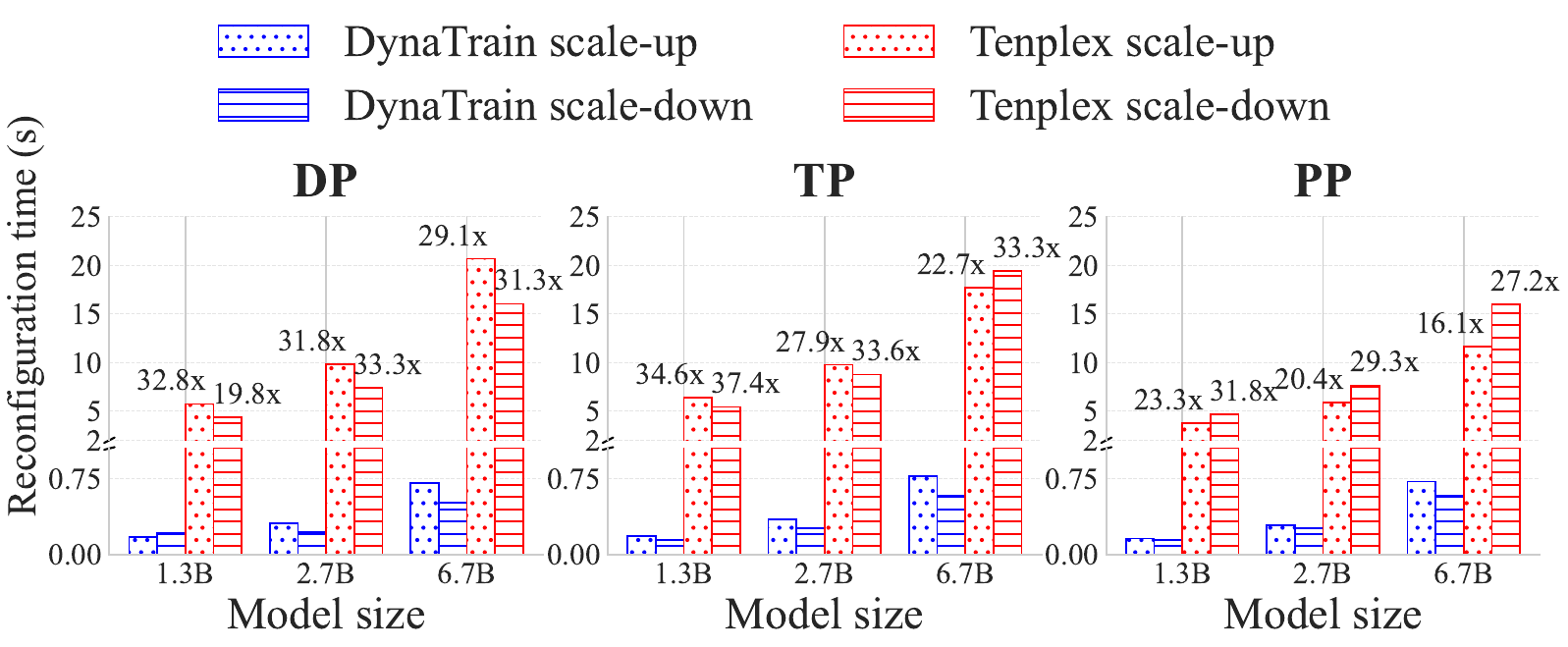}
    \caption{Single-dimension reconfiguration vs.\ Tenplex.}
    \label{fig:reconfig_tenplex}
\end{figure}

\vspace{-25pt}

\begin{figure}[ht]
    \centering
    \includegraphics[width=1.0\linewidth]{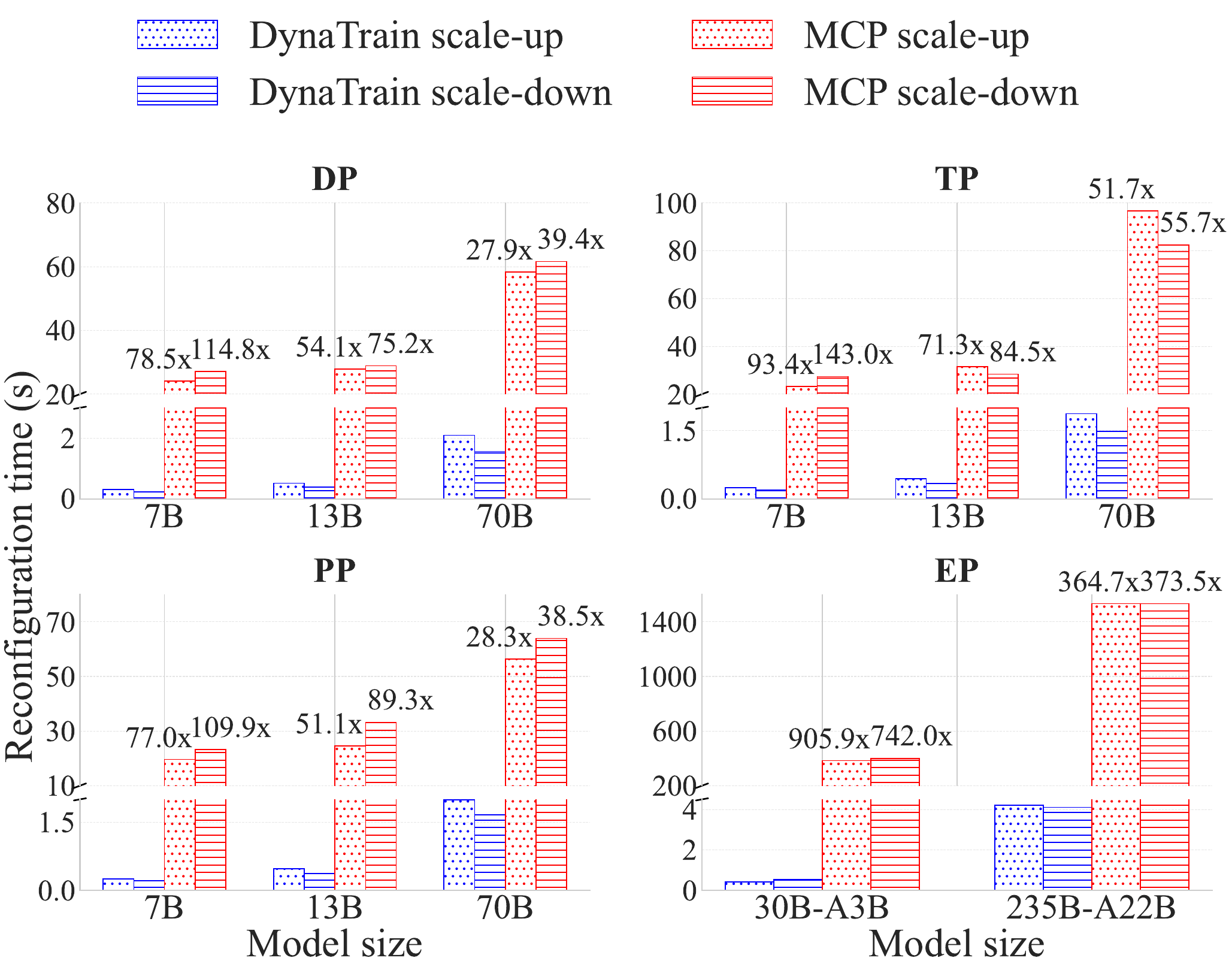}
    \caption{Single-dimension reconfiguration vs.\ MCP.}
    \vspace{-20pt}
    \label{fig:reconfig_single_dimension}
\end{figure}

\begin{table}[t]
\centering
\small
\setlength{\tabcolsep}{5pt}

\begin{tabular}{c|c|c|c}
\hline
\textbf{Model} & \textbf{Method} & \textbf{Src $\rightarrow$ Tgt} & \textbf{Time (s)} \\
\hline

\multirow[c]{4}{*}{\makecell{LLaMA\\ 7B}} 
& MCP
& \multirow{2}{*}{\makecell{
$\quad \ s_1(2,2,2)$ \\
$\to s_2(4,4,1)$
}}
& 14.63 + 15.78  \\

& \oursys
& 
& 0.818 \textbf{(37.18$\times$)}\\
\cline{2-4}

& MCP
& \multirow{2}{*}{\makecell{
$s_2\to s_1$
}}
& 11.18 + 19.37 \\

& \oursys
& 
& 0.874 \textbf{(34.95$\times$)}\\

\hline

\multirow[c]{4}{*}{\makecell{LLaMA\\ 13B}} 


& MCP
& \multirow{2}{*}{\makecell{
$\quad \ s_3(4,4,1)$ \\
$\to s_4(2,8,2)$
}}
& 16.76 + 13.26\\

& \oursys
& 
&  0.57 \textbf{(52.67$\times$)}\\
\cline{2-4}



& MCP
& \multirow{2}{*}{\makecell{
$s_4\to s_3$
}}
& 13.87 + 28.18 \\

& \oursys
& 
&  0.51 \textbf{(82.45$\times$)}\\

\hline

\multirow[c]{4}{*}{\makecell{LLaMA\\ 70B}} 
& MCP
& \multirow{2}{*}{\makecell{
$\quad \ s_5(4,8,2)$ \\
$\to s_6(8,16,1)$
}}
& 20.11 + 72.77 \\

& \oursys
& 
& 1.96 \textbf{(47.39$\times$)}\\
\cline{2-4}

& MCP
& \multirow{2}{*}{\makecell{
$s_6\to s_5$
}}
& 25.2 + 60.79 \\

& \oursys
& 
& 1.62 \textbf{(53.08$\times$)}\\

\hline
\hline

\multirow[c]{8}{*}{\makecell{LLaMA\\ 32B}} 
& HotSpa
& \multirow{2}{*}{\makecell{
$\quad \ s_7(4,4,2)$  \\
$\to s_8(2,8,2)$
}}
&  0.669 \\

& \oursys
& 
& 0.526 \textbf{(1.27$\times$)}\\
\cline{2-4}

& HotSpa
& \multirow{2}{*}{\makecell{
$s_8\to s_7$
}}
& 0.439 \\

& \oursys
& 
&  0.337 \textbf{(1.30$\times$)}\\
\cline{2-4}

& HotSpa
& \multirow{2}{*}{\makecell{
$\quad \ s_9(2,8,2)$ \\
$\to s_{10}(1,16,2)$
}}
& 0.826 \\

& \oursys
& 
& 0.392 \textbf{(2.11$\times$)}\\
\cline{2-4}

& HotSpa
& \multirow{2}{*}{\makecell{
$s_{10}\to s_9$
}}
&  0.253 \\

& \oursys
& 
&  0.299 \textbf{(0.85$\times$)}\\

\hline
\hline
\multirow[c]{4}{*}{\makecell{Qwen3 \\ 30B\\A3B}} 

& MCP
& \multirow{2}{*}{\makecell{
$\quad \ s_{11}(4,8,1,4)$\\
$\to s_{12}(2,4,4,2)$
}}
& 322.57 + 66.65  \\

& \oursys
& 
& 0.42 \textbf{(926.71$\times$)} \\
\cline{2-4}

& MCP
& \multirow{2}{*}{\makecell{
$s_{12}\to s_{11}$
}}
& 322.57 + 60.54 \\

& \oursys
& 
& 0.42 \textbf{(912.17$\times$)} \\

\hline

\multirow[c]{4}{*}{\makecell{Qwen3\\235B\\A22B}} 

& MCP
& \multirow{2}{*}{\makecell{
$\quad \ s_{13}(4,8,4,16)$ \\
$\to s_{14}(8,16,1,8)$
}}
& 1268.06 + 188.33 \\

& \oursys
& 
& 4.36 \textbf{(334.03$\times$)} \\
\cline{2-4}

& MCP
& \multirow{2}{*}{\makecell{
$s_{14} \to s_{13}$
}}
& 1289.89 + 229.32 \\

& \oursys
& 
& 4.36 \textbf{(348.44$\times$)} \\

\hline

\end{tabular}

\caption{Hybrid parallelism reconfiguration time across model scales. Dense models use $(\text{TP}, \text{PP}, \text{DP})$, while MoE models add EP. MCP time includes both saving and loading phases.}
\label{tab:hybrid_merged}
\vspace{-15pt}
\end{table}

\subsection{Parallelism Reconfiguration}


We evaluate parallelism reconfiguration across DP, TP, PP, and EP on LLaMA-2, GPT-3, and Qwen models, scaling from 8 to 128 GPUs. 

\subsubsection{Single-Dimension Reconfiguration}
\label{subsubsec:baseline}

\para{Comparison with Tenplex.} We first compare \oursys with Tenplex on GPT-3 models (1.3B, 2.7B, and 6.7B) using a 16-GPU cluster. Starting from $(\text{TP}, \text{PP}, \text{DP}) = (4, 2, 1)$, we independently double and halve each dimension to measure scale-up and scale-down overheads. \autoref{fig:reconfig_tenplex} reports the bidirectional reconfiguration times for DP, TP, and PP. \oursys consistently outperforms Tenplex, reaching up to 32.8$\times$ and 33.3$\times$ speedups for DP scale-up and scale-down. The gap stems from a fundamental difference in design: Tenplex relies on full-state disk serialization, whereas \oursys redistributes tensors directly GPU-to-GPU in memory.

Among the three dimensions, TP shows the highest speedup (up to 37.4$\times$ for the 1.3B model). Because Tenplex is checkpoint-based, its overhead grows near-linearly with model size regardless of dimension. \oursys benefits especially in TP because TP ranks are typically co-located within a node and can use high-bandwidth intra-node links such as NVLink. DP and PP require more inter-node communication, yielding slightly lower but still substantial speedups.

\para{Comparison with MCP.} We next evaluate larger-scale and EP reconfigurations against MCP. At 128 GPUs, we measure DP, TP, and PP reconfiguration on LLaMA-2 models (7B, 13B, and 70B), starting from $(4, 4, 4)$. \autoref{fig:reconfig_single_dimension} shows that \oursys achieves even larger gains at scale, outperforming MCP by 114.8$\times$, 143.0$\times$, and 109.9$\times$ for DP, TP, and PP, respectively. These gains come from direct in-memory P2P transfers, whose aggregate bandwidth grows with cluster size. Even on 128 GPUs, \oursys keeps 70B reconfiguration under 2\,s. As in the 16-GPU case, TP benefits most from fast intra-node links, while DP and PP are more constrained by inter-node bandwidth.


We also evaluate EP reconfiguration on Qwen-30B-A3B and Qwen-235B-A22B, starting from $(\text{TP}, \text{PP}, \text{DP}, \text{EP})=(4, 8, 4, 8)$. EP yields the largest gains, reaching up to 906$\times$ on the 30B MoE model. MCP suffers heavily because it checkpoints all parameters, including large numbers of fragmented expert tensors. In contrast, \oursys moves only the shards affected by the configuration change.

\subsubsection{Hybrid-Dimension Reconfiguration}
\label{dense_model}
Building on the single-dimension results, we now evaluate \oursys on more complex hybrid-dimension transitions, where multiple parallel dimensions change simultaneously.

\para{Hybrid Reconfiguration on Dense Models.} The first panel of ~\autoref{tab:hybrid_merged} reports reconfiguration times for LLaMA-2 across cluster scales from 2 to 16 nodes and model sizes from 7B to 70B.

\oursys consistently outperforms MCP. For 7B, it completes reconfiguration in under 0.9\,s, yielding $34.95\times$ to $37.18\times$ speedups over MCP, whose two-phase save-then-load path takes over 30\,s. For 13B, \oursys finishes in 0.51--0.57\,s, delivering $52.67\times$ to $82.45\times$ speedups. Even for 70B on 16 nodes, \oursys finishes in 1.62--1.96\,s, versus MCP's 85.99\,s and 92.88\,s, for $47.39\times$ and $53.08\times$ speedups.

Two trends emerge as model size grows. First, MCP's overhead scales super-linearly, reflecting increasing checkpoint I/O pressure. Second, \oursys remains in the sub-2\,s regime even at 70B, showing that in-memory remapping adds only modest coordination overhead. MCP is fundamentally constrained by checkpoint I/O and load-time tensor reconstruction, whereas \oursys avoids both through in-memory remapping and reduced data movement.

To compare against an in-memory baseline, we evaluate \oursys against HotSpa on a LLaMA-32B model (middle panel of ~\autoref{tab:hybrid_merged}). HotSpa relies on pre-compiled static execution graphs and supports only parameter-level intra-step changes. Even when we restrict \oursys to match this setup, it remains comparable or better, completing transitions in 0.30$\sim$0.53\,s. In three of four cases, \oursys outperforms HotSpa by 1.27$\times$ to 2.11$\times$; in the remaining case, the gap is only 0.046\,s. This shows that \oursys preserves the flexibility of dynamic reconfiguration without sacrificing the performance of specialized systems.



\para{Hybrid Reconfiguration on MoE Models.} The bottom panel of \autoref{tab:hybrid_merged} reports hybrid TP/PP/DP/EP transitions on Qwen3 MoE models at 4-node and 16-node scales. For Qwen3-30B-A3B, \oursys completes bidirectional reconfiguration between $(\text{TP}, \text{PP}, \text{DP}, \text{EP})$ = $(4, 8, 1, 4)$ and $(2, 4, 4, 2)$ in 0.42\,s in both directions, whereas MCP requires 389.2\,s and 383.1\,s, yielding 926.7$\times$ and 912.2$\times$ speedups. For Qwen3-235B-A22B, \oursys reconfigures between $(4, 8, 4, 16)$ and $(8, 16, 1, 8)$ in 4.36\,s, while MCP takes 1456.4\,s and 1519.2\,s, corresponding to 334.0$\times$ and 348.4$\times$ speedups.

These speedups are even larger than in dense models because MoE models contain many fragmented expert tensors, which cause severe I/O thrashing in MCP during serialization. \oursys avoids this penalty by batching expert communication and moving only the affected state shards. As model scale increases, \oursys's overhead rises modestly with in-memory redistribution, whereas MCP's overhead scales sharply with checkpoint size.



\vspace{-10pt}

\subsection{Elastic Resource Scaling}

We evaluate \oursys's efficiency in two elastic resource scaling scenarios: \textit{task migration} (transferring a training task to an entirely new set of devices) and \textit{dynamic cluster scaling} (handling node joins and departures during active training).

\subsubsection{Task Migration}
\begin{figure}[t]
    \centering
    \includegraphics[width=0.6\linewidth]{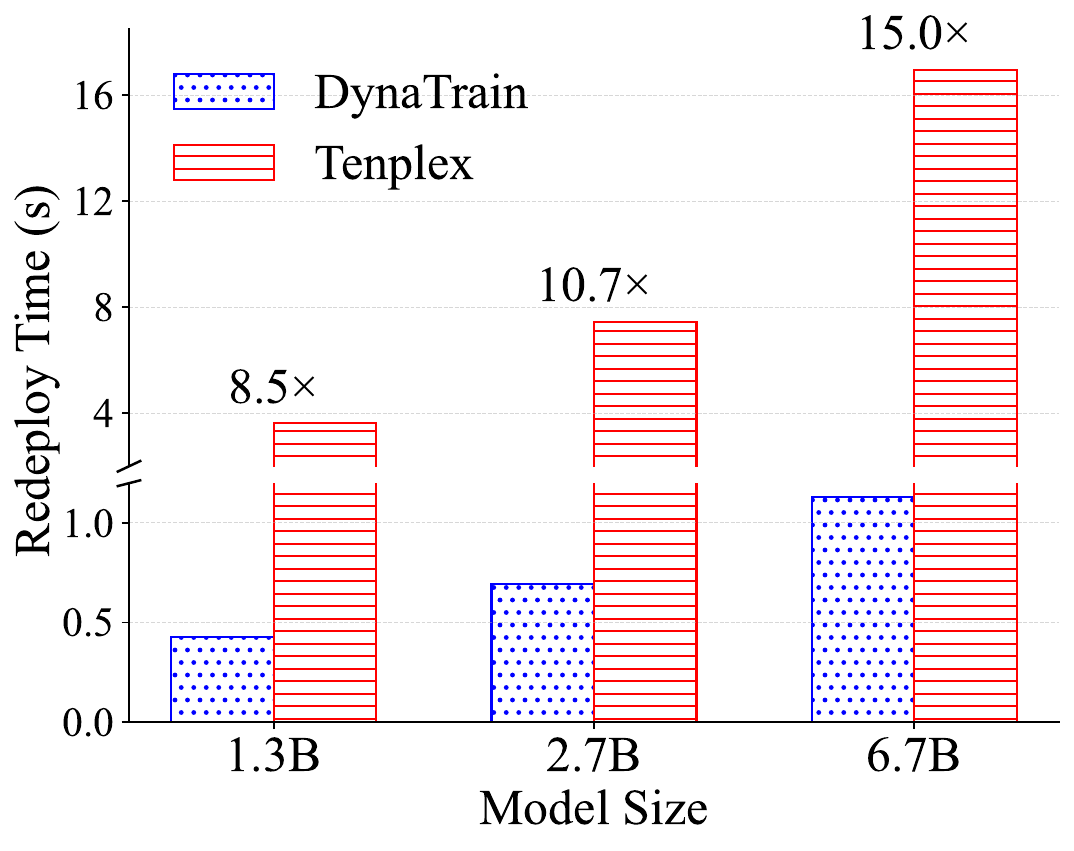}
    \caption{Job migration time compared with Tenplex.}
    \vspace{-15pt}
    \label{fig:redeploy}
\end{figure}

We first measure the time to migrate a training task to a new set of GPUs. To ensure a fair comparison, we migrate GPT-3 models across two 8-GPU nodes while keeping the parallelism configuration unchanged at $(\text{TP}, \text{PP}, \text{DP})$ = $(4, 1, 2)$. 

As shown in \autoref{fig:redeploy}, \oursys completes the migration of 1.3B, 2.7B, and 6.7B parameter models in just 0.43\,s, 0.69\,s, and 1.13\,s, respectively. This yields significant speedups of 8.5$\times$, 10.7$\times$, and 15.0$\times$ over Tenplex. As noted in Section~\ref{subsubsec:baseline}, \oursys's in-memory state transfer bypasses the costly checkpoint serialization required by Tenplex. This architectural advantage is particularly pronounced in the migration setting: Tenplex must route the training state through shared persistent storage. In contrast, \oursys utilizes direct peer-to-peer network transfers, completely eliminating intermediate I/O. Consequently, as the model size grows, Tenplex's I/O bottleneck becomes more severe, driving \oursys's speedup from 8.5$\times$ at 1.3B to 15.0$\times$ at 6.7B parameters.

\subsubsection{Dynamic Cluster Scaling}




We next evaluate dynamic cluster scaling using a LLaMA-2 13B model on a 32-GPU A800 cluster, simulating both node addition and node removal events. \autoref{tab:inter_scaling} breaks down the reconfiguration costs for the two resource-change scenarios. 

During node addition, \oursys's asynchronous initialization effectively hides 40.98\,s of process group setup overhead by overlapping it with ongoing computation. This leaves a mere 1.91\,s of exposed training interruption (incurred by state transfer and resharding), yielding a 95.6\% overlap ratio. Node removal exhibits similar efficiency: the system masks 40.32s of initialization time, exposing only 3.05\,s of interruption (a 93.0\% overlap ratio).

Overall, these results demonstrate that \oursys's overlap mechanism successfully masks the dominant cost of process group re-initialization, reducing the perceived training interruption during dynamic resource changes.

\begin{table}[t]
\centering
\begin{tabular}{lccc}
\toprule
\textbf{Scenario} & \textbf{Train. Intr.} & \textbf{Overlap (s)} & \textbf{Ratio} \\
\midrule
Addition   & 1.91 & 40.98 & 95.6\% \\
Removal & 3.05 & 40.32 & 93.0\% \\
\bottomrule
\end{tabular}
\caption{Latency breakdown under dynamic cluster scaling.}
\label{tab:inter_scaling}
\end{table}

\begin{table}[ht]
\centering
\begin{tabular}{cccc}
\toprule
\textbf{Transition} & \textbf{Naive} & \textbf{Buffer + Sync} & \textbf{Buffer + Async} \\
\midrule
$s_1 \to s_2$ & 3.24 & 1.27 (\textbf{2.55}$\times$) & 0.67 (\textbf{4.84}$\times$) \\
$s_2 \to s_1$ & 3.43 & 1.25 (\textbf{2.74}$\times$) & 0.64 (\textbf{5.36}$\times$) \\
\bottomrule
\end{tabular}
\caption{Ablation study of communication optimizations. Transition times are reported in seconds. $s_1$: $(\text{TP}, \text{PP}, \text{DP})$ = $(2, 8, 1)$ and $s_2 = (2, 2, 4)$.}
\label{tab_ablation}
\vspace{-15pt}
\end{table}

\subsection{Ablation Study}
\label{subsubsec_ablation}

To understand the sources of performance gains, we conduct an ablation study on LLaMA-2 13B at the 2-node scale (16 GPUs), isolating the effects of our tensor buffering and asynchronous communication mechanisms in the state transition engine. \autoref{tab_ablation} breaks down the reconfiguration time for transitions between two parallel strategies $s_1$ and $s_2$. 

\para{Impact of Tensor Buffering.} The baseline naive implementation executes synchronous unbuffered point to point transfers. This approach requires 3.24 s and 3.43 s for the two transition directions. By introducing tensor buffering, \oursys consolidates fragmented small tensor transfers into larger contiguous memory blocks. This severely reduces kernel launch overheads and improves network bandwidth utilization. As shown in the table, buffering alone drops the reconfiguration time to 1.27 s and 1.25 s. This delivers up to a 2.74$\times$ speedup over the naive baseline.

\para{Impact of Asynchronous Execution.} While buffering improves throughput, synchronous execution still forces the system to block on network transfers. By enabling asynchronous communication, \oursys overlaps the data transfers of different tensor chunks, effectively hiding network latency. This optimization further halves the latency, bringing the reconfiguration time down to 0.67 s and 0.64 s. 

Combined, buffering and asynchronous execution yield an overall 4.84$\times$ and 5.36$\times$ speedup for the respective directions, successfully compressing a penalty of multiple seconds into an operation taking well under one second.

\begin{figure}[ht]
    \centering
    \includegraphics[width=0.7\linewidth]{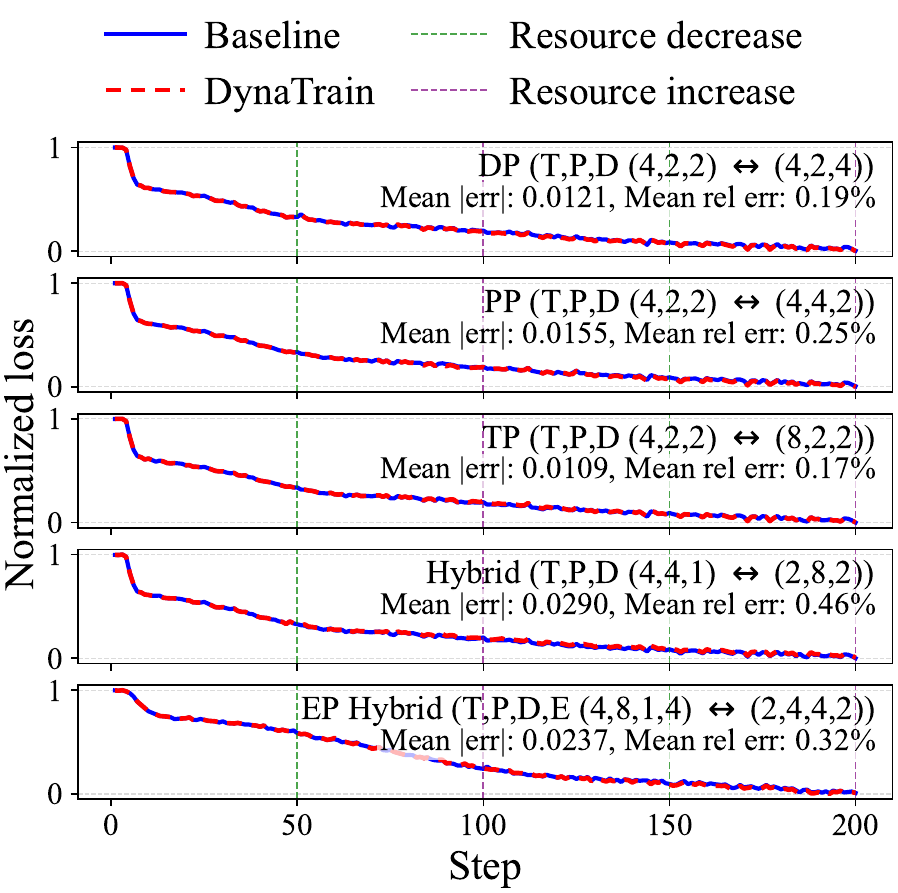}
    \caption{Convergence under elastic resharding.}
    \label{fig:validation}
    \vspace{-15pt}
\end{figure}

\subsection{Correctness Validation via Convergence}
\label{subsec:convergence}
We validate the numerical correctness of \oursys{}'s resharding mechanism by comparing its loss convergence trajectories against a static baseline (without reconfiguration) training a LLaMA-2 13B model on 32 GPUs. As depicted in \autoref{fig:validation}, we evaluate single-dimension scaling along DP, PP, and TP for dense models, as well as complex hybrid reconfigurations that simultaneously alter all dimensions for both dense and MoE models. To rigorously test stability, we alternate \oursys{} between the paired configurations every 50 steps, simulating periodic resource increases and decreases. Across all scenarios, the loss trajectory of \oursys{} perfectly tracks the baseline without any divergence or unexpected spikes. This confirms that \oursys{} executes exact state resharding while strictly maintaining hyperparameter consistency and global data ordering, ensuring that elastic scaling does not compromise model convergence.

\vspace{-10pt}
\section{Related Works}

\para{Checkpoint-Based Elastic Training.} Checkpoint-based systems adapt to reconfiguration by persisting training state and restarting execution. Megatron-LM Distributed Checkpointing (MCP)~\cite{megatron} provides a practical checkpointing foundation for large-scale training, while ByteCheckpoint~\cite{bytecheckpoint} improves flexibility through load-time reshaping across parallel configurations. Tenplex~\cite{wagenlander2024tenplex} further reduces unnecessary state movement through its Parallel Tensor Collection abstraction. Nevertheless, these approaches still route reconfiguration through stable storage and job restart, making latency fundamentally sensitive to checkpoint I/O and reload overhead.

\para{Checkpoint-Free Elastic Training.} Checkpoint-free designs avoid restart overhead but are typically more specialized. Bamboo~\cite{bamboo} and Oobleck~\cite{oobleck-sosp23} accelerate switching through redundancy or precomputed templates, but primarily target pipeline-parallel settings. HotSPa~\cite{hotspa} performs efficient in-memory switching within a training step, yet assumes a fixed device topology and pre-registered execution plans. In contrast, \oursys supports online reconfiguration across TP, PP, DP, and EP, while also handling changes in the allocated device set, enabling both parallelism reconfiguration and elastic resource scaling within one system.

\vspace{-10pt}
\section{Conclusion}

Large-scale LLM training is increasingly shaped by dynamic resources and multi-phase workflows, 
    yet today's training stacks still treat parallelism as a static decision made once at launch.
\oursys argues that parallelism hot switching should be the systems primitive for adapting training to such dynamics.
By decoupling distributed training states from their physical layouts through the Virtual Parameter Space, and by integrating routing, transition, and communication-topology management into one system, 
    \oursys makes online reconfiguration practical rather than exceptional. 
We view this as a step toward more adaptive training runtimes, 
    where parallelism is no longer fixed infrastructure, 
    but a flexible control dimension that can continuously respond to the needs of the workload and the cluster.





\bibliographystyle{plain}
\bibliography{reference}

\end{document}